%% file: main.tex
\documentclass[10pt,twocolumn,letterpaper]{article}

\usepackage[pagenumbers]{cvpr} 


\definecolor{cvprblue}{rgb}{0.21,0.49,0.74}
\usepackage[pagebackref,breaklinks,colorlinks,allcolors=cvprblue]{hyperref}

\usepackage[hypcap=false]{caption}
\usepackage{mdframed}

\usepackage{booktabs}
\usepackage{multirow}
\usepackage{xcolor}
\usepackage{colortbl}

\usepackage{epigraph}

\def\paperID{*****} 
\def\confName{****}
\def\confYear{2025}

\title{MCP-Zero: Active Tool Discovery for Autonomous LLM Agents}

\author{Xiang Fei\\
{\tt\small xiangf@stu.xmu.edu.cn}
\and
Xiawu Zheng\thanks{Corresponding author} \\
{\tt\small zhengxiawu@xmu.edu.cn}
\and
Hao Feng \\
{\tt\small haof@mail.ustc.edu.cn}
}

\makeatletter
\AtBeginDocument{
\let\@oldmaketitle\@maketitle
\renewcommand{\@maketitle}{\@oldmaketitle
  \centering
  \includegraphics[width=0.87\linewidth]{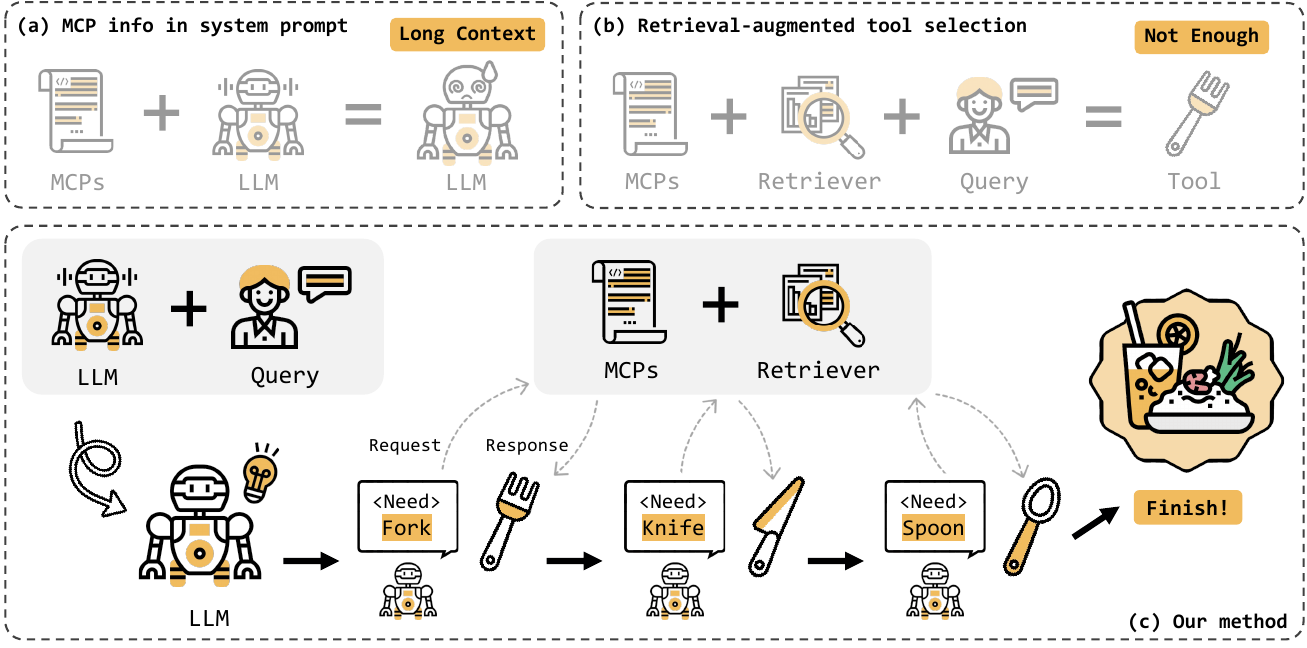}
  \captionof{figure}{
    Comparison of tool selection paradigms for LLM agents.
    (a) System-prompt-based methods inject all MCP tool schemas into the context, resulting in excessive prompt length and inefficiency.
    (b) Retrieval-augmented approaches select tools by matching the user query once, which may lead to inaccurate or insufficient tool selection, especially in multi-turn scenarios.
    (c) \textbf{MCP-Zero} enables the LLM to actively analyze the task, iteratively request the most relevant tools as needed, and dynamically construct a multi-step toolchain with minimal context overhead and high accuracy.
  }
  \label{fig:teaser}
  \vspace{17pt}
 }
}
\makeatother

\begin{document}

\maketitle

\begin{abstract}
    True intelligence requires active capability acquisition, yet current LLM agents inject pre-defined tool schemas into prompts, reducing models to passive selectors and falling short of robust general-purpose agency.
    We introduce MCP-Zero, an active agent framework that restores tool discovery autonomy to LLMs themselves. Instead of overwhelming models with all available tools, MCP-Zero enables agents to actively identify capability gaps, and request specific tools on-demand, transforming them from large-scale retrievers into genuine autonomous agents.
    The framework operates through three core mechanisms: (1) Active Tool Request, where models autonomously generate structured requests specifying their exact tool requirements; (2) Hierarchical Semantic Routing, a two-stage algorithm that matches requests to relevant servers and tools through improved semantic alignment; (3) Iterative Capability Extension, enabling agents to progressively build cross-domain toolchains while maintaining minimal context footprint.
    We construct MCP-tools, a comprehensive dataset of 308 MCP servers and 2,797 tools from the official Model-Context-Protocol repository. Experiments demonstrate that MCP-Zero preserves agent autonomy while achieving substantial efficiency gains: (i) accurate tool selection from nearly 3k candidates across 248.1k tokens; (ii) 98\% reduction in token consumption on APIBank while maintaining high accuracy; and (iii) consistent multi-turn performance that scales with tool ecosystem growth. This work establishes active tool discovery as a fundamental design pattern for scalable autonomous agent systems.
    The code and dataset is released at \url{https://github.com/xfey/MCP-Zero}.
\end{abstract}

\section{Introduction}




\epigraph{\textit{We only think when we are confronted with a problem.}}{\textit{John Dewey}}

True thinking emerges not from passive consumption but from active engagement with challenges. Intelligence manifests through the capacity to actively seek solutions rather than merely selecting from predetermined options.

Large language models (LLMs) have evolved rapidly, demonstrating increasingly sophisticated reasoning and problem-solving capabilities~\cite{yang2025qwen3,grattafiori2024llama,achiam2023gpt}. The emergence of function-calling mechanisms has enabled large language models (LLMs) to transcend their parametric boundaries, leveraging external tools, APIs, and execution environments to accomplish complex real-world tasks~\cite{qin2024tool,schick2023toolformer,masterman2024landscape,patil2024gorilla}.


However, current tool integration architectures fundamentally compromise agent autonomy. The dominant paradigm injects comprehensive tool schemas into system prompts, forcing agents into a passive role where they select from pre-defined options rather than actively discovering capabilities as needed (Figure~\ref{fig:teaser}a). This approach creates two critical problems: \textbf{massive context overhead} and \textbf{constrained decision autonomy}. For instance, the GitHub MCP server requires over 4,600 tokens for 26 tools, while comprehensive tool ecosystems can exceed 248k tokens—effectively transforming capable reasoning models into overwhelmed database query systems.


Recent retrieval-based approaches attempt to reduce context overhead by pre-selecting relevant tools based on semantic similarity with user queries~\cite{gan2025rag,moon2024efficient}. While these methods address the scaling problem, they perpetuate the fundamental issue of passive tool consumption. 
Query like ``Debug the file" requires filesystem access, code analysis, and command execution (Figure~\ref{fig:teaser}b): static retrieval based on initial queries cannot anticipate the evolving tool needs that emerge during task execution. More critically, this paradigm violates a core principle of autonomous agents—the ability to actively shape their environment based on dynamic assessment of their own capabilities.


The limitations of current approaches stem from three architectural constraints that prevent genuine agent autonomy: (1) external decision authority—tool selection is delegated to retrieval systems rather than the agent itself; (2) semantic distribution gaps—user queries and formal tool specifications exist in different semantic spaces, reducing matching precision; and (3) static capability assumptions—tools are selected once rather than discovered iteratively as task understanding evolves.

\begin{figure}[t]
    \centering
    \begin{mdframed}[linewidth=1pt, linecolor=gray!50, backgroundcolor=gray!5]
    \scriptsize
    \texttt{<function>\\\{"description": "Search for GitHub repositories", "name": "mcp\_github\_search\_repositories", "parameters": \{"\$schema": "http://json-schema.org/draft-07/schema\#", "additionalProperties": false, "properties": \{"page": \{"description": "Page number for pagination (default: 1)", "type": "number"\}, "perPage": \{"description": "Number of results per page (default: 30, max: 100)", "type": "number"\}, "query": \{"description": "Search query (see GitHub search syntax)", "type": "string"\}\}, "required": ["query"], "type": "object"\}\}\\</function>}
    \end{mdframed}
    \vspace{-10pt}
    \caption{Example of a single MCP tool definition from the GitHub MCP server. This tool requires 143 tokens, while the complete server requires over 4,600 tokens.}
    \label{fig:mcp_tool_example}
\end{figure}


\noindent \textbf{Toward Active Tool Discovery:} True autonomous agents must retain authority over their capability acquisition. Modern LLMs possess sophisticated reasoning, self-reflection, and planning capabilities that enable them to assess their own limitations and articulate specific tool requirements. Rather than constraining agents to pre-selected tool sets, we propose active tool discovery—a paradigm where agents autonomously identify capability gaps and request appropriate tools on-demand.
Based on this principle, we introduce MCP-Zero (Figure~\ref{fig:teaser}c), an active agent framework that restores tool discovery autonomy to LLMs through three core mechanisms:


\textbf{Active Tool Request.} Instead of passive tool consumption, agents generate structured requests that specify their exact requirements:

{
\scriptsize
\begin{quote}
\texttt{<tool\_assistant>}\\
\texttt{~~~server: ...~~\# Platform/permission domain}\\
\texttt{~~~tool: ...~~~~\# Operation type + target}\\
\texttt{</tool\_assistant>}
\end{quote}
}

This approach ensures semantic alignment between agent needs and tool documentation while preserving decision autonomy.



\textbf{Hierarchical Semantic Routing.} A two-stage matching algorithm first filters candidate servers by platform requirements, then ranks tools within selected servers based on semantic similarity. This hierarchical approach reduces search complexity while maintaining precision.

\textbf{Iterative Capability Extension.} Agents can discover and integrate tools throughout task execution, building cross-domain capabilities dynamically. When initial tools prove insufficient, agents can refine their requests and discover alternatives, providing natural fault tolerance.


To support systematic evaluation, we also construct MCP-tools, a comprehensive dataset comprising 308 servers and 2,797 tools from the official Model-Context-Protocol repository\footnote{\url{github.com/modelcontextprotocol/servers}}. Experiments demonstrate that MCP-Zero maintains agent autonomy while achieving substantial efficiency gains: 98\% reduction in token consumption with preserved accuracy across multi-turn conversations and large-scale tool ecosystems.

Our main contributions establish active tool discovery as a fundamental design pattern for autonomous agent systems:
\begin{itemize}
\item We propose MCP-Zero, enabling agents to maintain decision autonomy through active tool discovery, transforming them from passive selectors to autonomous capability architects.
\item We design hierarchical semantic routing that preserves semantic alignment between agent requests and tool specifications while reducing computational complexity.
\item We construct and release MCP-tools, providing the research community with a comprehensive evaluation framework for tool discovery systems.
\item We demonstrate that active discovery paradigms scale effectively with tool ecosystem growth while maintaining the autonomous decision-making that defines genuine agent systems.
\end{itemize}

\section{Related Work}

\subsection{Tool-Augmented LLMs}

The evolution of tool-augmented large language models has progressed through distinct paradigms, 
each addressing different aspects of the fundamental challenge: how to effectively integrate 
external capabilities with language model reasoning.

\textbf{Early Task-Specific Integration.} 
Initial approaches focused on hard-coding specific tools into language models. 
MRKL~\cite{karpas2022mrkl} introduced modular reasoning by combining neural networks with 
symbolic modules like calculators, while WebGPT~\cite{nakano2021webgpt} demonstrated web browsing 
capabilities for information retrieval tasks. Though effective within their domains, 
these systems suffered from limited scalability and poor generalization to new tool types.

\textbf{Universal Agent Protocols.} 
The introduction of ReAct~\cite{yao2023react} marked a paradigm shift by establishing 
the ``observation-action-thought" pattern, creating a universal protocol for 
tool-augmented reasoning. This framework became the foundation for modern agent systems 
including LangChain Agents~\cite{langchain2022} and AutoGen~\cite{wu2023autogen}, 
enabling flexible and extensible tool integration architectures.

\textbf{Training-Based Tool Learning.} 
Parallel developments explored learning tool usage through model training. 
Toolformer~\cite{schick2023toolformer} pioneered self-supervised learning for API call generation, 
teaching models to insert tool invocation markers within natural text generation. 
Gorilla~\cite{patil2024gorilla} extended this approach by constructing large-scale instruction 
datasets with tool calls, enabling supervised learning of query-to-tool mappings across 
diverse API collections.

\textbf{Context-Based Tool Injection.} 
The emergence of ChatGPT Function Calling and systems like HuggingGPT~\cite{shen2023hugginggpt} 
introduced context-based approaches, which inject JSON-Schema tool descriptions into 
system prompts, eliminating the need for specialized training. ART~\cite{paranjape2023art} 
further refined this paradigm by constructing demonstration libraries in chain-of-thought format, 
leveraging in-context learning for tool selection and usage.

\textbf{Fundamental Limitations.} 
Despite these advances, existing approaches face critical scalability challenges. 
Training-based methods require expensive retraining for each toolset update, limiting 
their adaptability to evolving tool ecosystems. Context-based methods, while more flexible, 
suffer from prohibitive context overhead when injecting comprehensive tool descriptions—a 
problem that becomes acute as tool collections scale to thousands of APIs. 
Most critically, when relevant tools are absent from the predefined context, 
task completion becomes impossible, highlighting the need for dynamic, on-demand 
tool discovery mechanisms that can adapt to diverse and evolving task requirements.

\subsection{Tool Retrieval for LLMs}

The success of Retrieval-Augmented Generation (RAG) in addressing knowledge limitations through 
the ``retrieve-insert-generate" paradigm has inspired its adaptation to tool selection for LLMs~\cite{lewis2020retrieval}. 
Classical RAG frameworks like REALM~\cite{guu2020retrieval}, RETRO~\cite{borgeaud2022improving} and 
In-Context RALM~\cite{ram2023context} demonstrated the effectiveness of dynamically incorporating 
external knowledge into generation processes. Recent work has extended this paradigm to tool 
selection, aiming to reduce context overhead by retrieving only the most relevant tools for 
a given query.

\textbf{Semantic Similarity-Based Retrieval.} 
Early approaches focused on direct semantic matching between user queries and tool descriptions. 
Gorilla~\cite{patil2024gorilla} constructed vector databases from API documentation and usage examples, 
employing semantic similarity to retrieve relevant tools. Tool2Vec~\cite{moon2024efficient} addressed the 
semantic gap between user requests and formal API descriptions by pre-collecting diverse user 
invocation patterns and computing averaged embeddings, though this approach requires extensive 
user interaction datasets for training. RAG-MCP~\cite{gan2025rag} performs server-level matching 
between user queries and documentation from MCPBench~\cite{luo2025evaluation}, returning all tools from 
the most similar server as the context of LLMs.

\textbf{Hierarchical Tool Retrieval.} 
Recognizing the limitations of flat retrieval, several works have explored coarse-to-fine 
approaches. AnyTool~\cite{du2024anytool} implemented a multi-level retrieval system based on the 
RapidAPI dataset, constructing separate retrievers for ``category-tool-API" hierarchies. 
ToolRerank~\cite{zheng2024toolrerank} leveraged pre-trained BERT models for semantic matching, 
while COLT~\cite{qu2024colt} employed specialized language models for tool selection. 
Re-Invoke~\cite{chen2024re} introduced key information extraction from user queries before 
tool matching.

\begin{figure}[t]
    \centering
    \includegraphics[width=0.94\columnwidth]{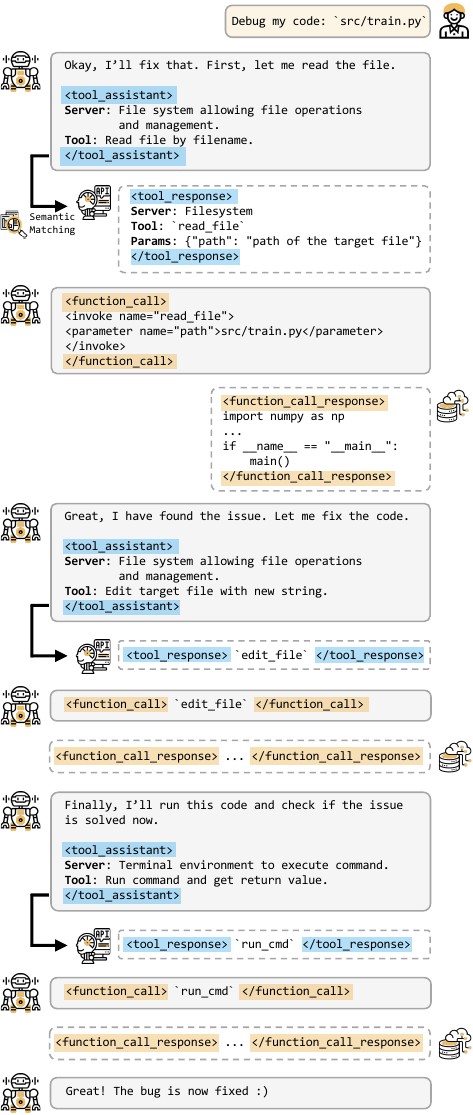}
    \caption{
        Illustrative example of MCP-Zero's iterative active invocation. The model progressively identifies capability gaps and requests tools across three domains on-demand.
    }
    \label{fig:main_workflow}
\end{figure}

\textbf{Limitations of Current Approaches.} 
Despite reducing context overhead, existing retrieval-based methods face fundamental limitations. 
First, they usually rely on single-round matching based on initial user queries: if the retrieval fails, 
task completion becomes impossible. Second, and more critically, 
user requests are often broad and composite in nature (e.g., ``write and run code to crawl AI 
repositories from GitHub"), requiring decomposition into multiple subtasks that cannot be solved 
by any single tool. Such scenarios demand progressive task breakdown where the model iteratively 
identifies capability gaps and requests appropriate tools on-demand, a challenge that current 
``query-once, retrieve-once" paradigms cannot adequately address.

\subsection{Model Context Protocol}

The Model Context Protocol (MCP) is an open standard introduced in 2024 to enable secure and 
uniform access to external tools and services for large language models~\cite{mcp2024}. 
By defining a standardized interface through JSON-RPC message exchange, MCP addresses the 
fragmentation problem in AI tool integration, where different platforms previously required 
custom connectors and proprietary protocols. This standardization has led to rapid adoption across major AI platforms and 
development environments, facilitating the creation of hundreds of MCP servers spanning 
diverse domains including file systems, databases, web services, and specialized APIs.

However, this expansion creates significant context overhead, as traditional approaches inject 
all server JSON-Schemas into system prompts simultaneously. Current solutions for MCP context 
optimization remain limited and follow conventional retrieval paradigms~\cite{gan2025rag}. 
Given the enhanced capabilities of modern LLMs and MCP's potential for agentic AI, we argue 
that existing methods fail to fully leverage contemporary language models' tool-calling 
capabilities, motivating our MCP-Zero framework for dynamic, on-demand tool discovery.

\section{Method: MCP-Zero}
\label{sec:mcp_zero}

We introduce MCP-Zero through two key aspects: the comprehensive framework design (Section~\ref{subsec:mcp_zero_framework}) and theoretical analysis of active retrieval (Section~\ref{subsec:mcp_zero_theoretical}). Our approach fundamentally shifts from passive tool injection to active tool discovery, enabling LLMs to dynamically construct task-specific toolchains across diverse domains with minimal context overhead.

\subsection{Active Tool Discovery Framework}
\label{subsec:mcp_zero_framework}

MCP-Zero is a active agent framework that enables LLMs to dynamically construct task-specific toolchains 
through on-demand tool retrieval. The framework operates 
through three core components that work synergistically to address the context overhead and multi-domain 
coordination challenges of existing approaches.

\textbf{Overall Workflow.} Given a user query such as ``Debug my code: \texttt{src/train.py}", the LLM analyzes 
the task and autonomously determines when external tool assistance is needed. As shown in 
Figure~\ref{fig:main_workflow}, the model progressively breaks down the complex debugging task into subtasks: 
reading the file, analyzing and fixing the code, and validating the fix through execution. At each step, 
instead of relying on pre-injected tool schemas, the model actively generates structured tool requests. 
These requests are processed through hierarchical vector routing to retrieve the most relevant tools, which 
are then injected into the context for immediate use. The model iteratively repeats this process throughout 
the conversation, constructing a cross-domain toolchain spanning filesystem operations, code editing, and 
command execution.

\textbf{Active Tool Request.} The foundation of our framework lies in returning tool requirement 
specification authority to the LLM itself. When the model identifies a capability gap that requires external 
assistance, it generates a structured request block:

{
\scriptsize
\begin{quote}
\texttt{<tool\_assistant>}\\
\texttt{~~~server: ...~~\# Platform/permission domain}\\
\texttt{~~~tool: ...~~~~\# Operation type + target}\\
\texttt{</tool\_assistant>}
\end{quote}
}
This mechanism enables the model to express tool needs spontaneously as they arise during task execution. 
Compared to raw user queries, the model-generated requests ensures better semantic alignment with tool documentation.
The \texttt{server} field specifies the platform or permission domain requirements, 
while the \texttt{tool} field describes the desired operation type and target. 

We design our framework around these two fields because they align naturally with the MCP specification, 
which mandates that all servers and tools provide descriptive documentation. 
This inherent requirement ensures consistent semantic information availability across the entire MCP ecosystem, 
making our retrieval approach universally applicable without additional metadata engineering.

Importantly, the model can generate multiple such requests throughout a single conversation, 
with each request triggering an independent retrieval process.

\textbf{Hierarchical Vector Routing.} To efficiently locate relevant tools from thousands of candidates, 
we employ a two-stage coarse-to-fine retrieval algorithm, using OpenAI \texttt{text-embedding-3-large} embeddings for semantic similarity matching. 

The system first filters candidate MCP servers by matching the \texttt{server} field against server descriptions. 
Since server descriptions are typically brief single sentences, we construct extended summaries that include 
comprehensive usage examples essential for accurate matching (detailed in Section~\ref{sec:mcp_tools}). 
We then perform matching against both the original descriptions and these enhanced summaries, 
taking the higher similarity score between the two approaches.
This dual-matching strategy leverages both original MCP documentation and enhanced summaries to improve retrieval precision. 

Subsequently, tools within selected servers are ranked based on semantic similarity between the \texttt{tool} field and tool descriptions. 
The final ranking score combines both server-level and tool-level similarities:
\begin{equation}
\text{score} = (s_{\text{server}} \times s_{\text{tool}}) \times \max(s_{\text{server}}, s_{\text{tool}})
\end{equation}
where $s_{\text{server}}$ and $s_{\text{tool}}$ represent cosine similarities at server and tool levels respectively. 
This scoring mechanism ensures that high similarity in either dimension contributes significantly to the final ranking, 
enabling effective recall of highly relevant tools. The system returns the top-$k$ tools, with dynamic 
adjustment possible when similarity scores are closely clustered. 
In our experiments, we achieve high accuracy with top-$1$ retrieval, though we can configure larger $k$ values to enhance fault tolerance when needed.

\textbf{Iterative Active Invocation.} Unlike traditional single-round retrieval approaches, 
MCP-Zero supports iterative tool discovery throughout task execution. After receiving retrieved tools, 
the model autonomously evaluates their adequacy for the current subtask. If the returned tools are insufficient 
or inappropriate, the model can refine its request specification and reinitiate retrieval, providing natural 
fault tolerance and self-correction capabilities. 
This iterative process continues until the model determines that either (1) suitable tools have been found and 
the task can proceed, or (2) no appropriate tools exist and the task should rely on the model's parametric knowledge.

The framework's key advantage lies in its ability to construct cross-domain toolchains dynamically. For complex 
tasks requiring coordination across multiple domains (e.g., filesystem access, code generation, and command execution in Figure~\ref{fig:main_workflow}), 
the model can progressively identify and request tools from different servers as subtask requirements become clear, 
avoiding the context overhead of pre-loading comprehensive tool collections while maintaining high task completion accuracy.
MCP-Zero represents a fundamental shift from ``predefined toolset" to ``dynamic on-demand tool discovery" for the community.

\subsection{Theoretical Analysis}
\label{subsec:mcp_zero_theoretical}

We provide a theoretical analysis of MCP-Zero's advantages over traditional approaches through formal modeling that connects to established theories in active learning and information acquisition.

\textbf{Problem Formulation.} Let $T = \{t_1, t_2, ..., t_n\}$ denote the complete tool collection, $q$ represent the user query, $s_t$ the current conversation state, and $t^*$ the optimal tool selection. Traditional approaches require simultaneous evaluation over the entire collection:
\begin{equation}
P_{\text{passive}}(t^*|q, T) = \frac{P(q|t^*, T)P(t^*|T)}{\sum_{t_i \in T} P(q|t_i, T)P(t_i|T)}
\end{equation}


MCP-Zero employs active information acquisition where agents generate requests $r$ based on their current state and capability assessment:
\begin{equation}
P_{\text{active}}(t^*|s_t) = \sum_{r} P(t^*|r)P(r|s_t)
\end{equation}
where $P(r|s_t)$ represents the agent's ability to articulate its needs given current understanding.

\textbf{Active Information Acquisition.} Tool request generation can be modeled as an active learning process where agents select actions to maximize information gain about task completion:
\begin{equation}
\begin{aligned}
r^* &= \arg\max_r I(T^*; r|s_t) \\
&= \arg\max_r [H(T^*|s_t) - H(T^*|r,s_t)]
\end{aligned}
\end{equation}
where $I(T^*; r|s_t)$ represents the mutual information between the optimal tool set and the request. This formulation captures the essence of active discovery: agents actively reduce uncertainty about their tool requirements.

\textbf{Scalability Analysis.} Traditional methods face fundamental scalability barriers:

\textit{Search Space Complexity:} Passive approaches must process all $n$ tools with $O(n)$ complexity. Active approaches first filter among $m$ servers ($m \ll n$), then match within filtered subsets, reducing complexity to $O(m + k)$ where $k$ is the average tools per selected server.

\textit{Attention Distribution:} With finite cognitive resources, passive methods distribute attention as $\frac{1}{n}$ per tool, degraded by noise factor $\eta(n) \propto \log(n)$. Active methods concentrate attention on relevant subsets, maintaining effectiveness as $\frac{1}{k}$ where $k \ll n$.

\textbf{Semantic Alignment Advantage.} The core advantage of active requests lies in improved semantic consistency. Agent-generated requests $r$ exhibit stronger alignment with tool documentation compared to raw user queries $q$:
\begin{equation}
\text{Alignment}(r,t) = \cos(\mathbf{e}_r, \mathbf{e}_t) > \cos(\mathbf{e}_q, \mathbf{e}_t)
\end{equation}
where $\mathbf{e}_r$, $\mathbf{e}_q$, $\mathbf{e}_t$ represent embeddings of request, query, and tool description respectively. This improvement stems from agents operating in the same semantic space as tool documentation.

\textbf{Iterative Information Gain.} Unlike single-shot retrieval, active discovery enables cumulative information acquisition over $k$ iterations:
\begin{equation}
I_{\text{total}} = \sum_{i=1}^k I(T^*; r_i|s_{i-1}) - \lambda \cdot \text{Cost}(r_i)
\end{equation}
where $\lambda$ represents the context overhead per request. This formulation captures the trade-off between information gain and computational efficiency that MCP-Zero optimizes.

In summary, the active paradigm provides measurable advantages:
\begin{itemize}
    \item Complexity Reduction: From $O(n)$ to $O(m + k)$ where $m + k \ll n$.
    \item Semantic Consistency: Direct embedding alignment in tool description space.
    \item Information Efficiency: Targeted uncertainty reduction rather than exhaustive search.
    \item Adaptive Capability: State-dependent tool discovery that evolves with task understanding.
\end{itemize}

These theoretical foundations explain the empirical performance gains observed in our experiments: 98\% token reduction with maintained accuracy reflects the efficiency of targeted information acquisition over exhaustive tool enumeration.

\section{Dataset: MCP-Tools}
\label{sec:mcp_tools}

To support our active tool discovery framework and enable comprehensive evaluation, we construct \textbf{MCP-tools}, the first retrieval-oriented dataset in the MCP domain. Unlike existing MCP evaluation frameworks such as MCPBench~\cite{luo2025evaluation} that focus on server availability and latency testing, our dataset is specifically designed to facilitate semantic tool discovery and retrieval for large language models.





\subsection{Dataset Construction}

Our dataset is constructed with the following steps:

\textbf{Data Collection.} We systematically collected MCP server information from the official Model Context Protocol repository\footnote{\url{github.com/modelcontextprotocol/servers}} (Tag: \texttt{2025.4.28}, Commit: \texttt{ad2d4e6}), encompassing 396 servers across three categories: 20 reference implementations, 114 third-party official servers, and 262 community contributions. This collection represents the current state of the MCP ecosystem and its diversity of tool capabilities.

\begin{figure}[t]
    \centering
    \begin{mdframed}[linewidth=1pt, linecolor=gray!50, backgroundcolor=gray!5]
    \scriptsize
    \begin{verbatim}
{
  "server_name": string,
  "server_description": string,
  "server_summary": string,
  "tools": [
    {
      "name": string,
      "description": string,
      "parameter": {
        "param1": "(type) description1",
        "param2": "(Optional, type) description2"
      }
    }
  ]
}
    \end{verbatim}
    \end{mdframed}
    \vspace{-10pt}
    \caption{The schema of the MCP-tools dataset.}
    \label{fig:mcp_tool_schema}
\end{figure}

\begin{figure*}[t]
    \centering
    \includegraphics[width=\textwidth]{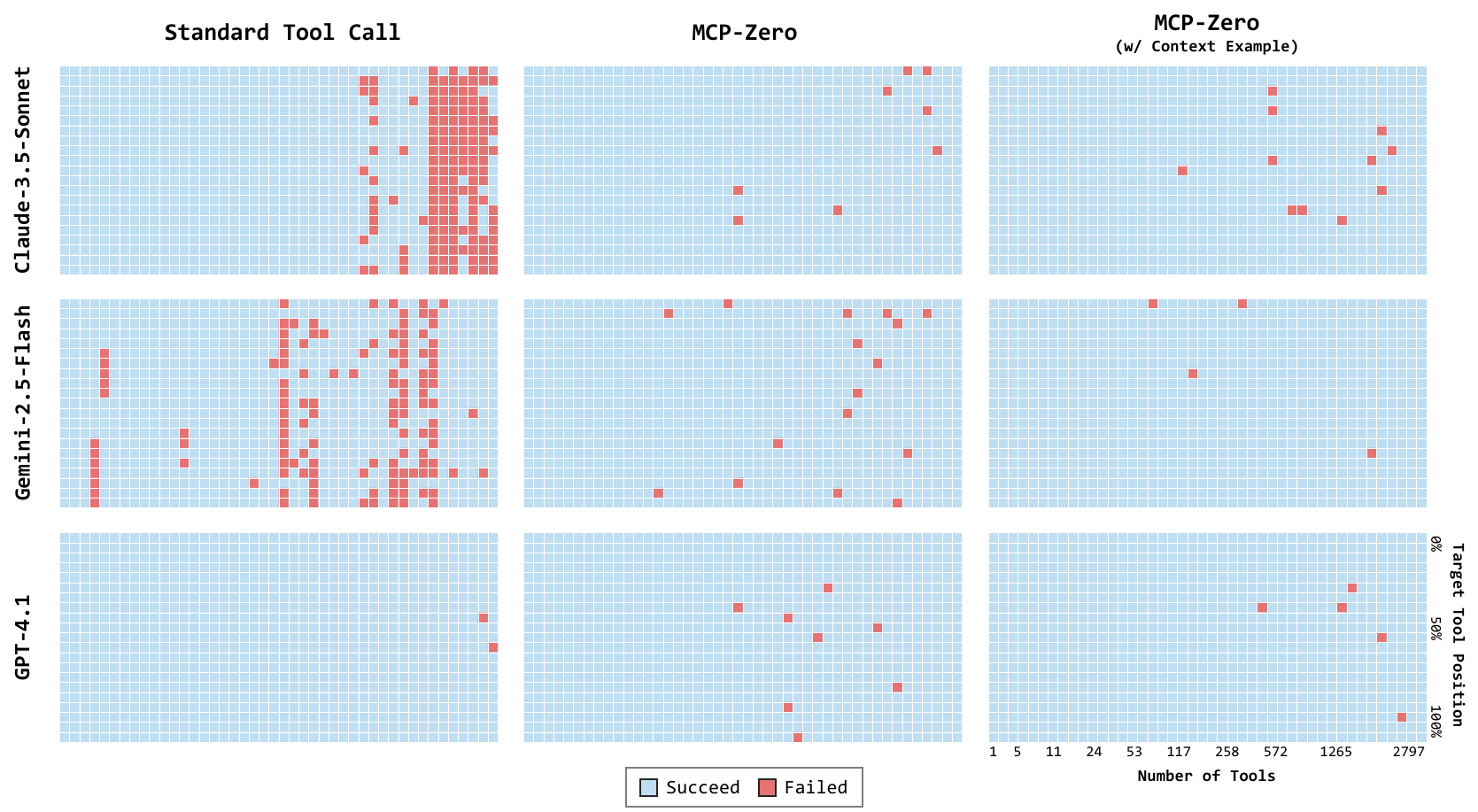}
    \caption{
        Needle-in-a-haystack test results demonstrating MCP-Zero's performance under extreme scale conditions with varying tool collection sizes. The left column shows baseline methods with standard tool call schemas; the middle column shows our MCP-Zero approach; the right column shows MCP-Zero enhanced with one ICL example to guide the model's description generation. Our method shows significant gains on Claude-3.5-Sonnet and Gemini-2.5-Flash, while GPT-4.1 shows no improvement due to its already strong baseline performance.
    }
    \label{fig:needle}
\end{figure*}

\textbf{Quality Assurance and Filtering.} We implemented rigorous filtering to ensure dataset completeness and semantic richness. Servers were retained only if they provided: (1) MCP-compliant tool definitions, (2) comprehensive documentation enabling semantic understanding, and (3) sufficient detail for meaningful retrieval evaluation. This process yielded 308 high-quality servers representing the core of the MCP ecosystem.




\textbf{Structured Information Extraction.} We employed Qwen2.5-72B-Instruct~\cite{qwen2025qwen25technicalreport} with carefully designed few-shot examples to extract structured information following a standardized schema, which is demonstrated in Figure~\ref{fig:mcp_tool_schema}.
The \texttt{server\_summary} field represents a key innovation: LLM-generated comprehensive summaries that distill server capabilities while excluding operational details. This design enables effective semantic matching across diverse query formulations, from formal API specifications to natural language capability descriptions.


\subsection{Dataset Characteristics}




After the above process, we have completed the construction of the dataset as follows:

\textbf{Scale and Diversity.} The final dataset comprises 308 servers and 2,797 tools, providing comprehensive coverage of the MCP ecosystem. Tool distribution shows significant variance (mean: 9.08, median: 5.0, $\sigma$: 11.40), reflecting the ecosystem's diversity from lightweight utilities to comprehensive API suites. Over half the servers (162) contain $\le$5 tools, while specialized servers include 60+ tools.

\textbf{Semantic Infrastructure.} To support efficient retrieval evaluation, we pre-compute embeddings for all textual content using OpenAI \texttt{text-embedding-3-large}, creating a searchable vector index that enables rapid semantic matching. These embeddings are included in the dataset release to ensure reproducible evaluation.

\textbf{Research Impact.} MCP-tools represents the first dataset specifically designed for semantic tool discovery in the MCP domain, complementing performance-focused frameworks like MCPBench~\cite{luo2025evaluation}. The dataset's design supports not only our active discovery framework, but provides a general foundation for evaluating retrieval-based tool selection approaches across the research community.

\section{Experiments}
\label{sec:experiments}

\subsection{Existing Datasets and Their Limitations}
\label{subsec:existing_datasets}

Before presenting our experimental evaluation, we analyze existing tool-calling datasets and explain why they are insufficient for evaluating our MCP-Zero framework.

\textbf{API-Bank}~\cite{li2023api} provides tool information with multi-turn conversations and human-annotated test sets, making it the most suitable existing dataset for our evaluation. \textbf{ToolBench}~\cite{qin2023toolllm} collected 16,464 REST APIs from RapidAPI Hub but many APIs lack essential descriptions. \textbf{ToolBank}~\cite{moon2024efficient} improved upon ToolBench but relies on model-generated data. \textbf{ToolAlpaca}~\cite{tang2023toolalpaca} synthesized 3,938 instances from 400 APIs but targets scenarios mismatched with contemporary use cases. The \textbf{APIBench1} from Gorilla~\cite{patil2024gorilla} contains 1,645 APIs but uses entirely GPT-synthesized conversations. Another \textbf{APIBench2}~\cite{peng2022revisiting} is proposed, but its application scenarios have low relevance to MCP tool calls.

Existing datasets have limitations for evaluating MCP-Zero: API scenarios differ from contemporary MCP use cases, lack hierarchical server-tool organization, or miss critical evaluation fields. Therefore, we use API-Bank as a reference and create the MCP-tools dataset for comprehensive evaluation. We are continuing to investigate other available datasets and conducting further validation.

\begin{figure}[t]
    \centering
    \includegraphics[width=\columnwidth]{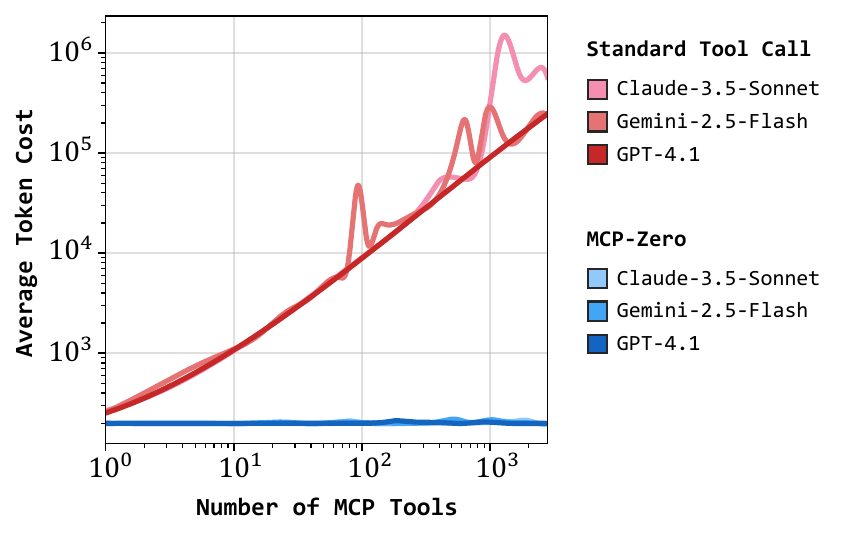}
    \caption{
        Token efficiency comparison in needle-in-a-haystack experiments. The graph shows the average token cost per successful retrieval across different collection sizes.
    }
    \label{fig:efficiency}
\end{figure}

\subsection{Needle-in-a-Haystack Evaluation}
\label{subsec:needle_evaluation}

To evaluate MCP-Zero's ability to accurately retrieve tools from large-scale collections under extreme context conditions, we conduct needle-in-a-haystack experiments on our MCP-tools dataset.

\textbf{Experimental Setup.} We construct test scenarios by injecting 1 to 2,797 tools into the environment, selecting task descriptions from various positions as queries, and requiring models to retrieve the target tool. This setup simulates the challenging scenario where relevant tools are buried within massive tool collections.
We compare three approaches:
\begin{itemize}
    \item \textbf{Baseline}: Standard tool call schemas with all tools injected into context
    \item \textbf{MCP-Zero}: Our active retrieval approach
    \item \textbf{MCP-Zero + ICL}: MCP-Zero enhanced with one in-context learning example to guide description generation
\end{itemize}

\begin{table}[t]
    \centering
    \resizebox{\columnwidth}{!}{
    \begin{tabular}{l|l|c|c|c|c}
    \toprule
    \textbf{Collection} & \textbf{Method} & \textbf{Claude-3.5} & \textbf{GPT-4.1} & \textbf{Gemini-2.5} & \textbf{Avg.\ Tokens} $\downarrow$ \\
    \midrule
    \multicolumn{6}{c}{\textit{Single-turn Conversation}} \\
    \midrule
    \multirow{3}{*}{Domain} 
    & Q.Retrieval & \multicolumn{3}{c|}{\cellcolor{gray!20}71.63} & {\cellcolor{gray!20}--} \\
    & Standard   & \textbf{97.60} & \textbf{98.08} & 92.79 & 312.4 \\
    & MCP-Zero   & 96.15 & 96.62 & \textbf{97.12} & \textbf{111.0} \textcolor{green!60!black}{(-64.47\%)} \\
    \midrule
    \multirow{3}{*}{Full} 
    & Q.Retrieval & \multicolumn{3}{c|}{\cellcolor{gray!20}71.63} & {\cellcolor{gray!20}--} \\
    & Standard   & 69.23 & 94.71 & 94.23 & 6308.2 \\
    & MCP-Zero   & \textbf{95.19} & \textbf{95.19} & \textbf{96.63} & \textbf{111.0} \textcolor{green!60!black}{(-98.24\%)} \\
    \midrule
    \multicolumn{6}{c}{\textit{Multi-turn Conversation}} \\
    \midrule
    \multirow{3}{*}{Domain} 
    & Q.Retrieval & \multicolumn{3}{c|}{\cellcolor{gray!20}65.05} & {\cellcolor{gray!20}--} \\
    & Standard   & \textbf{100.00} & \textbf{99.46} & 91.40 & 406.4 \\
    & MCP-Zero   & 91.40 & 93.01 & \textbf{93.01} & \textbf{159.0} \textcolor{green!60!black}{(-60.84\%)} \\
    \midrule
    \multirow{3}{*}{Full} 
    & Q.Retrieval & \multicolumn{3}{c|}{\cellcolor{gray!20}65.05} & {\cellcolor{gray!20}--} \\
    & Standard   & 60.22 & \textbf{93.01} & 92.47 & 6402.2 \\
    & MCP-Zero   & \textbf{90.32} & 92.47 & \textbf{94.62} & \textbf{159.0}  \textcolor{green!60!black}{(-97.52\%)}\\
    \bottomrule
    \end{tabular}}
    \caption{APIBank evaluation results comparing MCP-Zero with standard tool calling methods across different scenarios. The accuracy is based on the top-$1$ result. Results show accuracy (\%) for 
    three LLMs and average token consumption. ``Q.Retriaval" indicates retrieval based on user query.}
    \label{tab:apibank_results}
\end{table}

\textbf{Results Analysis.} As shown in Figure~\ref{fig:needle}, our method demonstrates significant performance gains on Claude-3.5-Sonnet and Gemini-2.5-Flash, while GPT-4.1 shows no improvement due to its already strong baseline performance across all tool collection sizes. Figure~\ref{fig:efficiency} illustrates the dramatic difference in token consumption between traditional approaches and MCP-Zero. While standard tool call methods exhibit exponential growth in token costs as the number of MCP tools increases, MCP-Zero maintains consistently low token usage.

\subsection{APIBank Evaluation}
\label{subsec:apibank_evaluation}

To validate MCP-Zero's effectiveness in realistic conversational tool retrieval scenarios, we conduct comprehensive experiments on the APIBank dataset. 

\textbf{Experimental Setup.} We extract description information from 
the APIBank level-1 dataset for retrieval tasks, get 48 unique tools in total, and process it for our evaluation 
framework. Since APIBank organizes data by individual APIs without 
server-level hierarchy, we directly retrieve tools without the 
server filtering stage.

We evaluate across two key dimensions: conversation context and tool collection scope. For conversation context, we test both single-turn scenarios (one user query for one response) and multi-turn scenarios (extended conversations with multiple exchanges). For tool collection scope, we examine both domain collections (using a curated subset of tools relevant to the specific domain) and full collections (retrieving from the complete set of all available tools).

\textbf{Results Analysis.} As shown in Table~\ref{tab:apibank_results}, we can conlude the following findings:
\begin{itemize}
    \item \textbf{Extreme Context Efficiency.} 
          MCP-Zero cuts prompt length by \textbf{60–98\%} across all settings 
          (e.g.\ 111 vs.\ 6.3k tokens in the full single-turn case), 
          validating its ability to "pay for tools only when they are needed".
          
    \item \textbf{Robust Scalability.} 
          When moving from a hand-curated \emph{Domain} subset to the \emph{Full} tool pool  
          (40x more APIs), standard schema-injection accuracy on Claude-3.5 plummets 
          from 97.60 to 69.23 (single-turn) and 100.00 to 60.22 (multi-turn);  
          MCP-Zero instead keeps accuracy at 95.19 / 90.32 respectively, 
          demonstrating strong resilience to attention dilution.
          
    \item \textbf{Multi-turn Consistency.} 
          MCP-Zero maintains high accuracy over conversation rounds ($\leq$3\% drop from single-
          to multi-turn), whereas standard methods degrade sharply once the context 
          accumulates previous calls and larger tool sets.
          
    \item \textbf{Necessity of Active Requests.}
          Pure query-retrieval baselines stall at 65–72 \% accuracy, confirming that 
          letting the model \emph{author} semantically aligned requests is crucial.
\end{itemize}

Experiments on APIBank corroborate our claims: MCP-Zero delivers near-optimal
or superior tool-selection accuracy while slashing context usage by up to two orders of
magnitude, remaining robust in both single- and multi-turn conversations and under
massive tool-pool scaling. These results highlight active, iterative tool discovery
as a practical path toward scalable, cost-efficient agent systems.

\section{Conclusion}
This work establishes active tool discovery as a fundamental paradigm for autonomous agent systems, enabling models to maintain decision autonomy while addressing critical scalability challenges in tool-calling architectures. MCP-Zero demonstrates that shifting from passive tool consumption to agent-driven capability acquisition achieves substantial efficiency gains—98\% token reduction with preserved accuracy—while restoring the core principle of autonomous agency: the ability to assess limitations and actively acquire necessary resources. Our theoretical framework, empirical validation, and the MCP-tools dataset provide both the foundation and infrastructure for advancing autonomous agent architectures as tool ecosystems continue expanding exponentially.

\section{Discussion}
\label{sec:discussion}

In this section we reflect on how the MCP‐Zero paradigm can be adopted by other
researchers (\S\ref{subsec:cookbook}), analyse the surprisingly gain from a
single in–context example (\S\ref{subsec:icl}), and position MCP‐Zero with respect
to the contemporaneous \textit{Alita} system, outlining a promising path toward
self-improving agentic AI (\S\ref{subsec:alita}).

\subsection{Cookbook: Integrate MCP-Zero Into Agent}
\label{subsec:cookbook}

MCP-Zero is fundamentally a simple yet effective approach that we hope will benefit the broader MCP community. The core methodology distills into three straightforward steps: prompting models to actively request tools, maintaining a lightweight tool index with semantic descriptions, and leveraging the improved semantic alignment for high-precision retrieval. Below we provide a practical guide for integrating these ideas into existing agent frameworks.

\textbf{Step 1 – Prompting the LLM to \emph{ask} for tools.}  
Give the model an explicit “permission” to declare missing capabilities.  In
practice this is a \texttt{system} instruction such as:  

\begin{quote}
\scriptsize
\texttt{If the current task cannot be solved with your own knowledge, emit a <tool\_assistant> block specifying the server domain and the tool operation you require.}
\end{quote}

In addition, the output structure needs to be specified as we mentioned in Section~\ref{subsec:mcp_zero_framework}. This step aims to stimulate the model's ability to "actively" propose requirements.

\smallskip
\textbf{Step 2 – Curate a lightweight MCP-style tool index.}  

Firstly, choose a scope based on your needs: the entire MCP-tools collection, a vertical slice (e.g.\ databases only), or your in-house APIs. Then, for every server/tool:

\begin{itemize}
   \item extract the name and description from metadata; 
   \item optionally let a strong LLM generate an \emph{enhanced summary} that
     emphasises capabilities and usage patterns;  
   \item store all texts in a vector store with pre-computed
     embeddings such as \texttt{text-embedding-3-large}.  
\end{itemize}

\smallskip
\textbf{Step 3 – Marry model output and retrieval.}  

When the agent emits a \texttt{<tool\_assistant>} block:

\begin{itemize}
   \item Match the \texttt{server} field against server descriptions and summaries; take top-\(m\) candidates.
   \item Within each candidate server, rank tools by the \texttt{tool} field with the tool description embeddings.
   \item Feed the best (or top-\(k\)) JSON-schemas back to the LLM.
\end{itemize}

Because the request text is already semantically aligned with the documents,
retrieval precision is higher than “user query → API doc” matching, maintaining performances while significantly conserving context.

\subsection{Why Does a Single ICL Example Help?}
\label{subsec:icl}

In \S\ref{subsec:needle_evaluation} we observed that adding \textbf{one} in-context
example (“ICL-1”) helps lifting needle-in-haystack accuracy marginally.
We hypothesise two simple but potent effects:

\begin{enumerate}
    \item \textbf{Stylistic anchor.}  
          Our base prompt merely says “output the server and tool you need”, but
          gives no example of \emph{how} the sentence should look like.  
          The single in-context sample provides the writing style as the reference, helping the generated requests land much closer to the curated descriptions, thus semantic matching becomes easier.

    \item \textbf{Semantic grounding.}  
          The example also clarifies the \emph{meaning} of each field, helping the model understand the specific definitions of MCP server and tool, thereby limiting its expression scope.
          After seeing this, the model reliably emits phrases such as
          \texttt{filesystem\_read} instead of a vague “read the file”, sharply
          reducing semantic mismatch.
\end{enumerate}

In short, a tiny demonstration patch acts as a \emph{schema anchor};
future work could replace ICL with a short grammar-based decoder rule,
but the one-shot approach is free and highly effective.

\subsection{Synergy with Alita: Using \emph{and} Making Tools}
\label{subsec:alita}

Concurrently, \textbf{Alita}~\cite{qiu2025alita} proposes a united manager agent that
\emph{creates} its own toolchain:
it web-searches for code, clones GitHub repos, builds environments, and executes
the resulting programs to accomplish tasks. We were pleasantly surprised by the contribution of this article, and found that the two lines of work are complementary:

\begin{itemize}
    \item MCP-Zero: \emph{efficiently \underline{finds and invokes} existing tools}  
    \item Alita   : \emph{automatically \underline{builds} missing tools on-the-fly}
\end{itemize}

MCP-Zero and Alita address complementary halves of the same problem: the former maximises \emph{tool discovery} while the latter maximises \emph{tool creation}. When combined, they form a virtuous loop: an agent first actively discover tools from \textit{all} available resources; if none fits, it switches to Alita’s workflow to synthesize a new one, then registers the freshly built tool for the community. 
We believe such a pipeline is a compelling
direction toward self-evolving, cost-aware agentic AI systems.

\subsection{Future Work}

While MCP-Zero demonstrates significant improvements in tool retrieval efficiency and accuracy, several promising directions warrant further investigation:

\textbf{Enhanced Experimental Validation.} Future work should expand evaluation across diverse domains. We plan to conduct comprehensive experiments on additional datasets to validate generalizability.

\textbf{Advanced Matching Algorithms.} The current semantic similarity approach could be enhanced. We envision incorporating multi-modal descriptions (e.g., code examples, usage patterns, parameter schemas) into the retrieval process, and exploring usage co-occurrence patterns for improved contextual understanding.

\textbf{MCP Server Implementation.} A natural extension involves packaging MCP-Zero as a dedicated MCP server providing tool discovery services. This "meta-server" would expose standardized APIs for active tool retrieval, enabling seamless integration into existing MCP ecosystems and serving as a centralized discovery hub for distributed tool collections.

\textbf{Multi-Agent Orchestration.} MCP-Zero's active discovery approach could enable better multi-agent collaboration. Future work could investigate how different agents can automatically discover and share tools with each other, allowing them to work together more effectively on complex tasks that require diverse capabilities.

\vspace{30pt}

{
    \small
    \bibliographystyle{ieeenat_fullname}
    \bibliography{main}
}


\end{document}